\documentclass[12pt]{article}
\usepackage{times}
\usepackage{graphicx}
\usepackage{float}
\usepackage{subfigure}
\usepackage{amsmath}
\usepackage{amssymb}
\usepackage{amsthm}
\usepackage{amsfonts}
\usepackage{rotating}
\usepackage[ruled,vlined]{algorithm2e}
\usepackage{algorithmic}
\usepackage{multirow}
\usepackage{multicol}
\usepackage{blindtext}
\newtheorem{lemma}{Lemma}
\newtheorem{theorem}{Theorem}

\newtheorem{proposition}{Proposition}

\makeatletter

\newcommand{\Rmnum}[1]{\expandafter\@slowromancap\romannumeral #1@}
\makeatother

\title{Bilateral Random Projections}

\begin{document}

\maketitle

\begin{abstract}
Low-rank structure have been profoundly studied in data mining and machine learning. In this paper, we show a dense matrix $X$'s low-rank approximation can be rapidly built from its left and right random projections $Y_1=XA_1$ and $Y_2=X^TA_2$, or bilateral random projection (BRP). We then show power scheme can further improve the precision. The deterministic, average and deviation bounds of the proposed method and its power scheme modification are proved theoretically. The effectiveness and the efficiency of BRP based low-rank approximation is empirically verified on both artificial and real datasets.
\end{abstract}

\section{Introduction}

\noindent Recent researches about low-rank structure concentrate on developing fast approximation and building meaningful decompositions. Two appealing representatives are the randomized approximate matrix decomposition \cite{RandomSVD} and column selection \cite{AdaptiveSampling}. The former proves that a matrix can be well approximated by its projection to the column space of its random projections. This rank-revealing method provides a fast approximation of SVD/PCA. The latter proves that a column subset of a low-rank matrix can span its whole range.

In this paper, we consider the problem of fast low-rank approximation. Given $r$ bilateral random projections (BRP) of an $m\times n$ dense matrix $X$ (w.l.o.g, $m\geq n$), i.e., $Y_1=XA_1$ and $Y_2=X^TA_2$, wherein $A_1\in\mathbb R^{n\times r}$ and $A_2\in\mathbb R^{m\times r}$ are random matrices,
\begin{equation}\label{E:lr_app}
L=Y_1\left(A_2^TY_1\right)^{-1}Y_2^T
\end{equation}
is a fast rank-$r$ approximation of $X$. The computation of $L$ includes an inverse of an $r\times r$ matrix and three matrix multiplications. Thus, for a dense $X$, $2mnr$ floating-point operations (flops) are required to obtain BRP, $r^2(2n+r)+mnr$ flops are required to obtain $L$. The computational cost is much less than SVD based approximation. The $L$ in (\ref{E:lr_app}) has been proposed in \cite{CandesBRP} as a recovery of a rank-$r$ matrix $X$ from $Y_1$ and $Y_2$, where $A_1$ and $A_2$ are independent Gaussian/SRFT random matrices. However, we propose that $L$ is a tight rank-$r$ approximation of a full rank matrix $X$, when $A_1$ and $A_2$ are correlated random matrices updated from $Y_2$ and $Y_1$, respectively. We then apply power scheme \cite{PowerScheme} to $L$ for improving the approximation precision, especially when the eigenvalues of $X$ decay slowly.

Theoretically, we prove the deterministic bound, average bound and deviation bound of the approximation error in BRP based low-rank approximation and its power scheme modification. The results show the error of BRP based approximation is close to the error of SVD approximation under mild conditions. Comparing with randomized SVD in \cite{RandomSVD} that extracts the column space from unilateral random projections, the BRP based method estimates both column and row spaces from bilateral random projections.

We give an empirical study of BRP on both artificial data and face image dataset. The results show its effectiveness and efficiency in low-rank approximation and recovery.

\section{Bilateral random projections (BRP) based low-rank approximation}

We first introduce the bilateral random projections (BRP) based low-rank approximation and its power scheme modification. The approximation error bounds of these two methods are discussed at the end of this section.

\subsection{Low-rank approximation with closed form}

In order to improve the approximation precision of $L$ in (\ref{E:lr_app}) when $A_1$ and $A_2$ are standard Gaussian matrices, we use the obtained right random projection $Y_1$ to build a better left projection matrix $A_2$, and use $Y_2$ to build a better $A_1$. In particular, after $Y_1=XA_1$, we update $A_2=Y_1$ and calculate the left random projection $Y_2=X^TA_2$, then we update $A_1=Y_2$ and calculate the right random projection $Y_1=XA_1$. A better low-rank approximation $L$ will be obtained if the new $Y_1$ and $Y_2$ are applied to (\ref{E:lr_app}). This improvement requires additional flops of $mnr$ in BRP calculation.

\subsection{Power scheme modification}

When singular values of $X$ decay slowly, (\ref{E:lr_app}) may perform poorly. We design a modification for this situation based on the power scheme \cite{PowerScheme}. In the power scheme modification, we instead calculate the BRP of a matrix $\tilde X=(XX^T)^qX$, whose singular values decay faster than $X$. In particular, $\lambda_i(\tilde X)={\lambda_i(\tilde X)}^{2q+1}$. Both $X$ and $\tilde X$ share the same singular vectors. The BRP of $\tilde X$ is:
\begin{equation}
Y_1=\tilde XA_1, Y_2=\tilde X^TA_2.
\end{equation}
According to (\ref{E:lr_app}), the BRP based $r$ rank approximation of $\tilde X$ is:
\begin{equation}
\tilde L=Y_1\left(A_2^TY_1\right)^{-1}Y_2^T.
\end{equation}
In order to obtain the approximation of $X$ with rank $r$, we calculate the QR decomposition of $Y_1$ and $Y_2$, i.e.,
\begin{equation}
Y_1=Q_1R_1, Y_2=Q_2R_2.
\end{equation}
The low-rank approximation of $X$ is then given by:
\begin{equation}\label{E:mlr_app}
L=\left(\tilde L\right)^{\frac{1}{2q+1}}=Q_1\left[R_1\left(A_2^TY_1\right)^{-1}R_2^T\right]^{\frac{1}{2q+1}}Q_2^T.
\end{equation}
The power scheme modification (\ref{E:mlr_app}) requires an inverse of an $r\times r$ matrix, an SVD of an $r\times r$ matrix and five matrix multiplications. Therefore, for dense $X$, $2(2q+1)mnr$ flops are required to obtain BRP, $r^2(m+n)$ flops are required to obtain the QR decompositions, $2r^2(n+2r)+mnr$ flops are required to obtain $L$. The power scheme modification reduces the error of (\ref{E:lr_app}) by increasing $q$. When the random matrices $A_1$ and $A_2$ are built from $Y_1$ and $Y_2$, $mnr$ additional flops are required in the BRP calculation.

\section{Approximation error bounds}

We analyze the error bounds of the BRP based low-rank approximation (\ref{E:lr_app}) and its power scheme modification (\ref{E:mlr_app}).

The SVD of an $m\times n$ (w.l.o.g, $m\geq n$) matrix $X$ takes the form:
\begin{equation}
X=U\Lambda V^T=U_1\Lambda_1V_1^T+U_2\Lambda_2V_2^T,
\end{equation}
where $\Lambda_1$ is an $r\times r$ diagonal matrix which diagonal elements are the first largest $r$ singular values, $U_1$ and $V_1$ are the corresponding singular vectors, $\Lambda_2$, $U_2$ and $V_2$ forms the rest part of SVD. Assume that $r$ is the target rank, $A_1$ and $A_2$ have $r+p$ columns for oversampling. We consider the spectral norm of the approximation error $E$ for (\ref{E:lr_app}):
\begin{align}
\notag\|X-L\|&=\left\|X-Y_1\left(A_2^TY_1\right)^{-1}Y_2^T\right\|\\
&=\left\|\left[I-XA_1\left(A_2^TXA_1\right)^{-1}A_2^T\right]X\right\|.
\end{align}
The unitary invariance of the spectral norm leads to
\begin{align}\label{E:unblock}
\notag \|X-L\|=\left\|U^T\left[I-XA_1\left(A_2^TXA_1\right)^{-1}A_2^T\right]X\right\|\\
~~~~~~~~~~=\left\|\Lambda\left[I-V^TA_1\left(A_2^TXA_1\right)^{-1}A_2^TU\Lambda\right]\right\|.
\end{align}

In low-rank approximation, the left random projection matrix $A_2$ is built from the left random projection $Y_1=XA_1$, and then the right random projection matrix $A_1$ is built from the left random projection $Y_2=X^TA_2$. Thus $A_2=Y_1=XA_1=U\Lambda V^TA_1$ and $A_1=Y_2=X^TA_2=X^TXA_1=V\Lambda^2V^TA_1$. Hence the approximation error given in (\ref{E:unblock}) has the following form:
\begin{equation}\label{E:newunblock}
\left\|\Lambda\left[I-\Lambda^2V^TA_1\left(A_1^TV\Lambda^4V^TA_1\right)^{-1}A_1^TV\Lambda^2\right]\right\|.
\end{equation}

The following Theorem \ref{T:deterministicBound} gives the bound for the spectral norm of the deterministic error $\|X-L\|$.

\begin{theorem}\label{T:deterministicBound}
{\rm \textbf{(Deterministic error bound)}} Given an $m\times n\left(m\geq n\right)$ real matrix $X$ with singular value decomposition $X=U\Lambda V^T=U_1\Lambda_1 V_1^T+U_2\Lambda_2 V_2^T$, and chosen a target rank $r\leq n-1$ and an $n\times(r+p)$ ($p\geq2$) standard Gaussian matrix $A_1$, the BRP based low-rank approximation (\ref{E:lr_app}) approximates $X$ with the error upper bounded by
\begin{equation}
\notag\|X-L\|^2\leq\left\|\Lambda_2^2\left(V_2^TA_1\right)(V_1^TA_1)^\dagger\Lambda_1^{-1}\right\|^2+\left\|\Lambda_2\right\|^2.
\end{equation}
\end{theorem}

See Section \ref{S:proof} for the proof of Theorem \ref{T:deterministicBound}.

If the singular values of $X$ decay fast, the first term in the deterministic error bound will be very small. The last term is the rank-$r$ SVD approximation error. Therefore, the BRP based low-rank approximation (\ref{E:lr_app}) is nearly optimal.

\begin{theorem}\label{T:deterministicBoundPower}
{\rm \textbf{(Deterministic error bound, power scheme)}} Frame the hypotheses of Theorem \ref{T:deterministicBound}, the power scheme modification (\ref{E:mlr_app}) approximates $X$ with the error upper bounded by
\begin{align}
\notag\|X-L\|^2\leq&\left(\left\|\Lambda_2^{2(2q+1)}\left(V_2^TA_1\right)\left(V_1^TA_1\right)^\dagger\Lambda_1^{-(2q+1)}\right\|^2\right.\\
\notag &\left.+\left\|\Lambda_2^{2q+1}\right\|^2\right)^{1/(2q+1)}.
\end{align}
\end{theorem}

See Section \ref{S:proof} for the proof of Theorem \ref{T:deterministicBoundPower}.

If the singular values of $X$ decay slowly, the error produced by the power scheme modification (\ref{E:mlr_app}) is less than the BRP based low-rank approximation (\ref{E:lr_app}) and decreasing with the increasing of $q$.

The average error bound of BRP based low-rank approximation is obtained by analyzing the statistical properties of the random matrices that appear in the deterministic error bound in Theorem \ref{T:deterministicBound}.

\begin{theorem}\label{T:AverageErrorBound}
{\rm \textbf{(Average error bound)}} Frame the hypotheses of Theorem \ref{T:deterministicBound},
\begin{align}
\notag\mathbb E\|X-L\|\leq&\left(\sqrt{\frac{1}{p-1}\sum\limits_{i=1}^r\frac{\lambda_{r+1}^2}{\lambda_i^2}}+1\right)|\lambda_{r+1}|\\
\notag &+\frac{{\rm e}\sqrt{r+p}}{p}\sqrt{\sum\limits_{i=r+1}^n\frac{\lambda_i^2}{\lambda_r^2}}.
\end{align}
\end{theorem}

See Section \ref{S:proof} for the proof of Theorem \ref{T:AverageErrorBound}.

The average error bound will approach to the SVD approximation error $|\lambda_{r+1}|$ if $|\lambda_{r+1}|\ll|\lambda_{i:i=1,\cdots,r}|$ and $|\lambda_{r}|\gg|\lambda_{i:i=r+1,\cdots,n}|$.

The average error bound for the power scheme modification is then obtained from the result of Theorem \ref{T:AverageErrorBound}.

\begin{theorem}\label{T:AverageErrorBoundPower}
{\rm (\textbf{Average error bound, power scheme})} Frame the hypotheses of Theorem \ref{T:deterministicBound}, the power scheme modification (\ref{E:mlr_app}) approximates $X$ with the expected error upper bounded by
\begin{align}
\notag E\|X-L\|\leq&\left[\left(\sqrt{\frac{1}{p-1}\sum\limits_{i=1}^r\frac{\lambda_{r+1}^{2(2q+1)}}{\lambda_i^{2(2q+1)}}}+1\right)|\lambda_{r+1}^{2q+1}|\right.\\
\notag &\left.+\frac{{\rm e}\sqrt{r+p}}{p}\sqrt{\sum\limits_{i=r+1}^n\frac{\lambda_i^{2(2q+1)}}{\lambda_r^{2(2q+1)}}}\right]^{1/(2q+1)}.
\end{align}
\end{theorem}

See Section \ref{S:proof} for the proof of Theorem \ref{T:AverageErrorBoundPower}.

Compared the average error bounds of the BRP based low-rank approximation with its power scheme modification, the latter produces less error than the former, and the error can be further decreased by increasing $q$.

The deviation bound for the spectral norm of the approximation error can be obtained by analyzing the deviation bound of $\left\|\Lambda_2^2\left(V_2^TA_1\right)(V_1^TA_1)^\dagger\Lambda_1^{-1}\right\|$ in the deterministic error bound and by applying the concentration inequality for Lipschitz functions of a Gaussian matrix.

\begin{theorem}\label{T:DeviationErrorBound}
{\rm (\textbf{Deviation bound})} Frame the hypotheses of Theorem \ref{T:deterministicBound}. Assume that $p\geq4$. For all $u,t\geq1$, it holds that
\begin{align}
\notag \left\|X-L\right\|\leq&\left(1+t\sqrt{\frac{12r}{p}}\left(\sum\limits_{i=1}^r\lambda_i^{-1}\right)^{\frac{1}{2}}+\frac{{\rm e}\sqrt{r+p}}{p+1}\cdot\right.\\
\notag &\left.tu\lambda_r^{-1}\right)\lambda_{r+1}^2+\frac{{\rm e}\sqrt{r+p}}{p+1}\cdot t\lambda_r^{-1}\left(\sum\limits_{i=r+1}^n\lambda_i^2\right)^{\frac{1}{2}}.
\end{align}
except with probability ${\rm e}^{-u^2/2}+4t^{-p}+t^{-(p+1)}$.
\end{theorem}

See Section \ref{S:proof} for the proof of Theorem \ref{T:DeviationErrorBound}.

\section{Proofs of error bounds}
\label{S:proof}

\subsection{Proof of Theorem \ref{T:deterministicBound}}

The following lemma and propositions from \cite{RandomSVD} will be used in the proof.

\begin{lemma}\label{L:conjugate}
Suppose that $M\succeq0$. For every $A$, the matrix $A^TMA\succeq0$. In particular,
\begin{equation}
M\preceq N~~\Rightarrow~~A^TMA\preceq A^TNA.
\end{equation}
\end{lemma}

\begin{proposition}\label{P:Range}
Suppose ${\rm range}(N)\subset{\rm range}(M)$. Then, for each matrix $A$, it holds that $\|\mathcal P_NA\|\leq\|\mathcal P_MA\|$ and that $\|(I-\mathcal P_M)A\|\leq\|(I-\mathcal P_N)A\|$.
\end{proposition}

\begin{proposition}\label{P:inversePurtubation}
Suppose that $M\succeq0$. Then
\begin{equation}
I-\left(I+M\right)^{-1}\preceq M.
\end{equation}
\end{proposition}

\begin{proposition}\label{P:blockNorm}
We have $\|M\|\leq\|A\|+\|C\|$ for each partitioned positive semidefinite matrix
\begin{equation}
M=\left[
    \begin{array}{cc}
      A & B \\
      B^T & C \\
    \end{array}
  \right].
\end{equation}
\end{proposition}

The proof of Theorem \ref{T:deterministicBound} is given below.

\begin{proof}
Since an orthogonal projector projects a given matrix to the range (column space) of a matrix $M$ is defined as $\mathcal P_M=M(M^TM)^{-1}M^T$, the deterministic error (\ref{E:newunblock}) can be written as
\begin{equation}\label{E:errorproj}
\|E\|=\left\|\Lambda\left(I-\mathcal P_M\right)\right\|,~M=\Lambda^2V^TA_1.
\end{equation}

By applying Proposition \ref{P:Range} to the error (\ref{E:errorproj}), because ${\rm range}(M(V_1^TA_1)^\dagger\Lambda_1^{-2})\subset{\rm range}(M)$, we have
\begin{equation}\label{E:projNM}
\|E\|=\left\|\Lambda\left(I-\mathcal P_M\right)\right\|\leq\left\|\Lambda\left(I-\mathcal P_N\right)\right\|,
\end{equation}
where
\begin{equation}\label{E:IsubPN}
N=\left[
    \begin{array}{c}
      \Lambda_1^2V_1^TA_1 \\
      \Lambda_2^2V_2^TA_1 \\
    \end{array}
  \right]
(V_1^TA_1)^\dagger\Lambda_1^{-2}=
\left[
  \begin{array}{c}
    I \\
    H \\
  \end{array}
\right]
.
\end{equation}
Thus $\left(I-\mathcal P_N\right)$ can be written as
\begin{equation}\label{E:IsubPNBlock}
\notag I-\mathcal P_N=
\left[
  \begin{array}{cc}
    I-\left(I+H^TH\right)^{-1} & -\left(I+H^TH\right)^{-1}H^T \\
    -H\left(I+H^TH\right)^{-1} & I-H\left(I+H^TH\right)^{-1}H^T \\
  \end{array}
\right]
\end{equation}

For the top-left block in (\ref{E:IsubPNBlock}), Proposition \ref{P:inversePurtubation} leads to $I-\left(I+H^TH\right)^{-1}\preceq H^TH$. For the bottom-right block in (\ref{E:IsubPNBlock}), Lemma \ref{L:conjugate} leads to $I-H\left(I+H^TH\right)^{-1}H^T\preceq I$. Therefore,
\begin{equation}
\notag I-\mathcal P_N\preceq\left[
                       \begin{array}{cc}
                         H^TH & -\left(I+H^TH\right)^{-1}H^T \\
                         -H\left(I+H^TH\right)^{-1} & I \\
                       \end{array}
                     \right]
\end{equation}

By applying Lemma \ref{L:conjugate}, we have
\begin{align}
\notag&\Lambda\left(I-\mathcal P_N\right)\Lambda\preceq\\
\notag&\left[
                                                  \begin{array}{cc}
                                                    \Lambda_1^TH^TH\Lambda_1 & -\Lambda_1^T\left(I+H^TH\right)^{-1}H^T\Lambda_2 \\
                                                    -\Lambda_2^TH\left(I+H^TH\right)^{-1}\Lambda_1 & \Lambda_2^T\Lambda_2 \\
                                                  \end{array}
                                                \right]
\end{align}

According to Proposition \ref{P:blockNorm}, the spectral norm of $\Lambda(I-\mathcal P_N)$ is bounded by
\begin{align}\label{E:rawBound}
\notag&\left\|\Lambda\left(I-\mathcal P_N\right)\right\|^2=\left\|\Lambda\left(I-\mathcal P_N\right)\Lambda\right\|\\
&\leq\left\|\Lambda_2^2\left(V_2^TA_1\right)(V_1^TA_1)^\dagger\Lambda_1^{-1}\right\|^2+\left\|\Lambda_2\right\|^2.
\end{align}

By substituting (\ref{E:rawBound}) into (\ref{E:projNM}), we obtain the deterministic error bound. This completes the proof.
\end{proof}

\subsection{Proof of Theorem \ref{T:deterministicBoundPower}}

The following proposition from \cite{RandomSVD} will be used in the proof.

\begin{proposition}\label{P:PowerNorm}
Let $\mathcal P$ be an orthogonal projector, and let $A$ be a matrix. For each nonnegative $q$,
\begin{equation}
\|\mathcal PA\|\leq\left\|\mathcal P\left(AA^T\right)^qA\right\|^{1/\left(2q+1\right)}.
\end{equation}
\end{proposition}

The proof of Theorem \ref{T:deterministicBoundPower} is given below.

\begin{proof}
The power scheme modification (\ref{E:mlr_app}) applies the BRP based low-rank approximation (\ref{E:lr_app}) to $\tilde X=(XX^T)^qX=U\Lambda^{2q+1}V^T$ rather than $X$. In this case, the approximation error is
\begin{equation}
\|\tilde X-\tilde L\|=\left\|\Lambda^{2q+1}\left(I-\mathcal P_M\right)\right\|,~M=\Lambda^{2(2q+1)}V^TA_1.
\end{equation}
According to Theorem \ref{T:deterministicBound}, the error is upper bounded by
\begin{align}\label{E:unblockPower}
\notag&\left\|\tilde X-\tilde L\right\|^2\leq\\
&\left\|\Lambda_2^{2(2q+1)}\left(V_2^TA_1\right)(V_1^TA_1)^\dagger\Lambda_1^{-(2q+1)}\right\|^2+\left\|\Lambda_2^{2q+1}\right\|^2.
\end{align}
The deterministic error bound for the power scheme modification is obtained by applying Proposition \ref{P:PowerNorm} to (\ref{E:unblockPower}). This completes the proof.
\end{proof}

\subsection{Proof of Theorem \ref{T:AverageErrorBound}}

The following propositions from \cite{RandomSVD} will be used in the proof.

\begin{proposition}\label{P:SGT}
Fix matrices $S$, $T$, and draw a standard Gaussian matrix $G$. Then it holds that
\begin{equation}
\mathbb E\left\|SGT^T\right\|\leq\|S\|\|T\|_F+\|S\|_F\|T\|.
\end{equation}
\end{proposition}

\begin{proposition}\label{P:pesudoinvGaussian}
Draw an $r\times(r+p)$ standard Gaussian matrix $G$ with $p\geq 2$. Then it holds that
\begin{align}
\mathbb E\|G^\dagger\|_F^2=\frac{r}{p-1},
\mathbb E\|G^\dagger\|\leq\frac{{\rm e}\sqrt{r+p}}{p}.
\end{align}
\end{proposition}

The proof of Theorem \ref{T:AverageErrorBound} is given below.

\begin{proof}
The distribution of a standard Gaussian matrix is rotational invariant. Since 1) $A_1$ is a standard Gaussian matrix and 2) $V$ is an orthogonal matrix, $V^TA_1$ is a standard Gaussian matrix, and its disjoint submatrices $V_1^TA_1$ and $V_2^TA_1$ are standard Gaussian matrices as well.

Theorem \ref{T:deterministicBound} and the H\"{o}lder's inequality imply that
\begin{align}\label{E:EXsubL}
\notag \mathbb E\|X-L\|&\leq\mathbb E\left(\left\|\Lambda_2^2\left(V_2^TA_1\right)(V_1^TA_1)^\dagger\Lambda_1^{-1}\right\|^2+\|\Lambda_2\|^2\right)^{1/2}\\
&\leq\mathbb E\left\|\Lambda_2^2\left(V_2^TA_1\right)(V_1^TA_1)^\dagger\Lambda_1^{-1}\right\|+\|\Lambda_2\|.
\end{align}
We condition on $V_1^TA_1$ and apply Proposition \ref{P:SGT} to bound the expectation w.r.t. $V_2^TA_1$, i.e.,
\begin{align}\label{E:ExAA}
\notag &E\left\|\Lambda_2^2\left(V_2^TA_1\right)(V_1^TA_1)^\dagger\Lambda_1^{-1}\right\|\\
\notag &\leq\mathbb E\left(\left\|\Lambda_2^2\right\|\left\|(V_1^TA_1)^\dagger\Lambda_1^{-1}\right\|_F+\left\|\Lambda_2^2\right\|_F\left\|(V_1^TA_1)^\dagger\Lambda_1^{-1}\right\|\right)\\
\notag&\leq\left\|\Lambda_2^2\right\|\left(\mathbb E\left\|(V_1^TA_1)^\dagger\Lambda_1^{-1}\right\|_F^2\right)^{1/2}+\\
&\left\|\Lambda_2^2\right\|_F\cdot\mathbb E\left\|(V_1^TA_1)^\dagger\right\|\cdot\left\|\Lambda_1^{-1}\right\|.
\end{align}
The Frobenius norm of $(V_1^TA_1)^\dagger\Lambda_1^{-1}$ can be calculated as
\begin{align}
\notag \left\|(V_1^TA_1)^\dagger\Lambda_1^{-1}\right\|_F^2&={\rm trace}\left[\Lambda_1^{-1}\left((V_1^TA_1)^\dagger\right)^T(V_1^TA_1)^\dagger\Lambda_1^{-1}\right]\\
\notag&={\rm trace}\left[\left(\left(\Lambda_1V_1^TA_1\right)\left(\Lambda_1V_1^TA_1\right)^T\right)^{-1}\right].
\end{align}
Since 1) $V_1^TA_1$ is a standard Gaussian matrix and 2) $\Lambda_1$ is a diagonal matrix, each column of $\Lambda_1V_1^TA_1$ follows $r$-variate Gaussian distribution $\mathcal N_r(\textbf{0},\Lambda_1^2)$. Thus the random matrix $\left(\left(\Lambda_1V_1^TA_1\right)\left(\Lambda_1V_1^TA_1\right)^T\right)^{-1}$ follows the inverted Wishart distribution $\mathcal W^{-1}_r(\Lambda_1^{-2},r+p)$. According to the expectation of inverted Wishart distribution \cite{AspectsStats}, we have
\begin{align}
\notag &\mathbb E\left\|(V_1^TA_1)^\dagger\Lambda_1^{-1}\right\|_F^2\\
\notag &=\mathbb E~{\rm trace}\left[\left(\left(\Lambda_1V_1^TA_1\right)\left(\Lambda_1V_1^TA_1\right)^T\right)^{-1}\right]\\
\notag &={\rm trace}~\mathbb E\left[\left(\left(\Lambda_1V_1^TA_1\right)\left(\Lambda_1V_1^TA_1\right)^T\right)^{-1}\right]\\
&=\frac{1}{p-1}\sum\limits_{i=1}^r\lambda_i^{-2}.
\end{align}
We apply Proposition \ref{P:pesudoinvGaussian} to the standard Gaussian matrix $V_1^TA_1$ and obtain
\begin{equation}
\mathbb E\left\|(V_1^TA_1)^\dagger\right\|\leq\frac{{\rm e}\sqrt{r+p}}{p}.
\end{equation}
Therefore, (\ref{E:ExAA}) can be further derived as
\begin{align}\label{E:ExAAfinal}
\notag E&\left\|\Lambda_2^2\left(V_2^TA_1\right)(V_1^TA_1)^\dagger\Lambda_1^{-1}\right\|\\
\notag&\leq\lambda_{r+1}^2\cdot\sqrt{\frac{1}{p-1}\sum\limits_{i=1}^r\lambda_i^{-2}}+\sqrt{\sum\limits_{i=r+1}^n\lambda_i^2}\cdot\frac{{\rm e}\sqrt{r+p}}{p}\cdot|\lambda_r^{-1}|\\
&=|\lambda_{r+1}|\sqrt{\frac{1}{p-1}\sum\limits_{i=1}^r\frac{\lambda_{r+1}^2}{\lambda_i^2}}+\frac{{\rm e}\sqrt{r+p}}{p}\sqrt{\sum\limits_{i=r+1}^n\frac{\lambda_i^2}{\lambda_r^2}}.
\end{align}
By substituting (\ref{E:ExAAfinal}) into (\ref{E:EXsubL}), we obtain the average error bound
\begin{align}
\notag\mathbb E\|X-L\|\leq&\left(\sqrt{\frac{1}{p-1}\sum\limits_{i=1}^r\frac{\lambda_{r+1}^2}{\lambda_i^2}}+1\right)|\lambda_{r+1}|+\\
&\frac{{\rm e}\sqrt{r+p}}{p}\sqrt{\sum\limits_{i=r+1}^n\frac{\lambda_i^2}{\lambda_r^2}}.
\end{align}
This completes the proof.
\end{proof}

\subsection{Proof of Theorem \ref{T:AverageErrorBoundPower}}

The proof of Theorem \ref{T:AverageErrorBoundPower} is given below.

\begin{proof}
By using H\"{o}lder's inequality and Theorem \ref{T:deterministicBoundPower}, we have
\begin{align}\label{E:powerXsubL}
\notag\mathbb E\left\|X-L\right\|&\leq\left(\mathbb E\left\|X-L\right\|^{2q+1}\right)^{1/(2q+1)}\\
&\leq\left(\mathbb E\left\|\tilde X-\tilde L\right\|\right)^{1/(2q+1)}.
\end{align}
We apply Theorem \ref{T:AverageErrorBound} to $\tilde X$ and $\tilde L$ and obtain the bound of $\mathbb E\|\tilde X-\tilde L\|$, noting that $\lambda_i(\tilde X)=\lambda_i(X)^{2q+1}$.
\begin{align}\label{E:tildeXsubL}
\notag \mathbb E\left\|\tilde X-\tilde L\right\|=&\left(\sqrt{\frac{1}{p-1}\sum\limits_{i=1}^r\frac{\lambda_{r+1}^{2(2q+1)}}{\lambda_i^{2(2q+1)}}}+1\right)|\lambda_{r+1}^{2q+1}|+\\
&\frac{{\rm e}\sqrt{r+p}}{p}\sqrt{\sum\limits_{i=r+1}^n\frac{\lambda_i^{2(2q+1)}}{\lambda_r^{2(2q+1)}}}.
\end{align}
By substituting (\ref{E:tildeXsubL}) into (\ref{E:powerXsubL}), we obtain the average error bound of the power scheme modification shown in Theorem \ref{T:AverageErrorBoundPower}. This completes the proof.
\end{proof}

\subsection{Proof of Theorem \ref{T:DeviationErrorBound}}

The following propositions from \cite{RandomSVD} will be used in the proof.

\begin{proposition}\label{P:LipschitzConcentration}
Suppose that $h$ is a Lipschitz function on matrices:
\begin{equation}
\left|h(X)-h(Y)\right|\leq L\|X-F\|_F~~for~all~X,Y.
\end{equation}
Draw a standard Gaussian matrix $G$. Then
\begin{equation}
\Pr\left\{h(G)\geq\mathbb Eh(G)+Lt\right\}\leq{\rm e}^{-t^2/2}.
\end{equation}
\end{proposition}

\begin{proposition}\label{P:GaussianNormDeviation}
Let $G$ be a $r\times(r+p)$ standard Gaussian matrix where $p\geq4$. For all $t\geq1$,
\begin{align}
\notag&\Pr\left\{\left\|G^\dagger\right\|_F\geq\sqrt{\frac{12r}{p}}\cdot t\right\}\leq4t^{-p}~~{\rm and}\\
&\Pr\left\{\left\|G^\dagger\right\|\geq\frac{{\rm e}\sqrt{r+p}}{p+1}\cdot t\right\}\leq t^{-(p+1)}.
\end{align}
\end{proposition}

The proof of Theorem \ref{T:DeviationErrorBound} is given below.

\begin{proof}
According to the deterministic error bound in Theorem \ref{T:deterministicBound}, we study the deviation of $\left\|\Lambda_2^2\left(V_2^TA_1\right)\left(V_1^TA_1\right)^\dagger\Lambda_1^{-1}\right\|$. Consider the Lipschitz function $h(X)=\left\|\Lambda_2^2X\left(V_1^TA_1\right)^\dagger\Lambda_1^{-1}\right\|$, its Lipschitz constant $L$ can be estimated by using the triangle inequality:
\begin{align}
\notag &\left|h(X)-h(Y)\right| \leq\left\|\Lambda_2^2\left(X-Y\right)\left(V_1^TA_1\right)^\dagger\Lambda_1^{-1}\right\|\\
\notag &\leq\left\|\Lambda_2^2\right\|\left\|X-Y\right\|\left\|\left(V_1^TA_1\right)^\dagger\right\|\left\|\Lambda_1^{-1}\right\|\\
&\leq\left\|\Lambda_2^2\right\|\left\|\left(V_1^TA_1\right)^\dagger\right\|\left\|\Lambda_1^{-1}\right\|\left\|X-Y\right\|_F.
\end{align}
Hence the Lipschitz constant satisfies $L\leq\left\|\Lambda_2^2\right\|\left\|\left(V_1^TA_1\right)^\dagger\right\|\left\|\Lambda_1^{-1}\right\|$. We condition on $V_1^TA_1$ and then Proposition \ref{P:SGT} implies that
\begin{align}
\notag\mathbb E\left[h\left(V_2^TA_1\right)\left|\right.V_1^TA_1\right]\leq&\left\|\Lambda_2^2\right\|\left\|\left(V_1^TA_1\right)^\dagger\right\|_F\left\|\Lambda_1^{-1}\right\|_F+\\
\notag&\left\|\Lambda_2^2\right\|_F\left\|\left(V_1^TA_1\right)^\dagger\right\|\left\|\Lambda_1^{-1}\right\|.
\end{align}
We define an event $T$ as
\begin{align}
\notag&T=\left\{\left\|\left(V_1^TA_1\right)^\dagger\right\|_F\leq\sqrt{\frac{12r}{p}}\cdot t~~{\rm and}~~\right.\\
&\left.~~~~~~~~~\left\|\left(V_1^TA_1\right)^\dagger\right\|\leq\frac{{\rm e}\sqrt{r+p}}{p+1}\cdot t\right\}.
\end{align}
According to Proposition \ref{P:GaussianNormDeviation}, the event $T$ happens except with probability
\begin{equation}
\Pr\left\{\overline{T}\right\}\leq4t^{-p}+t^{-(p+1)}.
\end{equation}
Applying Proposition \ref{P:LipschitzConcentration} to the function $h\left(V_2^TA_1\right)$, given the event $T$, we have
\begin{align}
\notag\Pr&\left\{\left\|\Lambda_2^2\left(V_2^TA_1\right)\left(V_1^TA_1\right)^\dagger\Lambda_1^{-1}\right\|>\right.\\
\notag&\left\|\Lambda_2^2\right\|\left\|\left(V_1^TA_1\right)^\dagger\right\|_F\left\|\Lambda_1^{-1}\right\|_F+\\
\notag&{\left\|\Lambda_2^2\right\|_F\left\|\left(V_1^TA_1\right)^\dagger\right\|\left\|\Lambda_1^{-1}\right\|+}\\
&\left.{\left\|\Lambda_2^2\right\|\left\|\left(V_1^TA_1\right)^\dagger\right\|\left\|\Lambda_1^{-1}\right\|\cdot u}\mid T\right\}\leq{\rm e}^{-u^2/2}.
\end{align}
According to the definition of the event $T$ and the probability of $\overline{T}$, we obtain
\begin{align}
\notag\Pr&\left\{\left\|\Lambda_2^2\left(V_2^TA_1\right)\left(V_1^TA_1\right)^\dagger\Lambda_1^{-1}\right\|>\right.\\
\notag&\left\|\Lambda_2^2\right\|\left\|\Lambda_1^{-1}\right\|_F\sqrt{\frac{12r}{p}}\cdot t+\left\|\Lambda_2^2\right\|_F\left\|\Lambda_1^{-1}\right\|\frac{{\rm e}\sqrt{r+p}}{p+1}\cdot t\\
\notag&\left.+\left\|\Lambda_2^2\right\|\left\|\Lambda_1^{-1}\right\|\frac{{\rm e}\sqrt{r+p}}{p+1}\cdot tu\right\}\leq\\
\notag&{\rm e}^{-u^2/2}+4t^{-p}+t^{-(p+1)}.
\end{align}
Therefore,
\small
\begin{align}
\notag\Pr&\left\{\left\|\Lambda_2^2\left(V_2^TA_1\right)\left(V_1^TA_1\right)^\dagger\Lambda_1^{-1}\right\|+\left\|\Lambda_2\right\|>\right.\\
\notag&\left.\left(1+t\sqrt{\frac{12r}{p}}\left(\sum\limits_{i=1}^r\lambda_i^{-1}\right)^{1/2}+\frac{{\rm e}\sqrt{r+p}}{p+1}\cdot tu\lambda_r^{-1}\right)\lambda_{r+1}^2+\right.\\
\notag&\left.\frac{{\rm e}\sqrt{r+p}}{p+1}\cdot t\lambda_r^{-1}\left(\sum\limits_{i=r+1}^n\lambda_i^2\right)^{1/2}\right\}\leq\\
&{\rm e}^{-u^2/2}+4t^{-p}+t^{-(p+1)}.
\end{align}
\normalsize
Since Theorem \ref{T:deterministicBound} implies $\left\|X-L\right\|\leq\left\|\Lambda_2^2\left(V_2^TA_1\right)\left(V_1^TA_1\right)^\dagger\Lambda_1^{-1}\right\|+\left\|\Lambda_2\right\|$,
we obtain the deviation bound in Theorem \ref{T:DeviationErrorBound}. This completes the proof.
\end{proof}

\section{Empirical Study}

\begin{figure}[htb]
\begin{center}
 \includegraphics[width=1\linewidth]{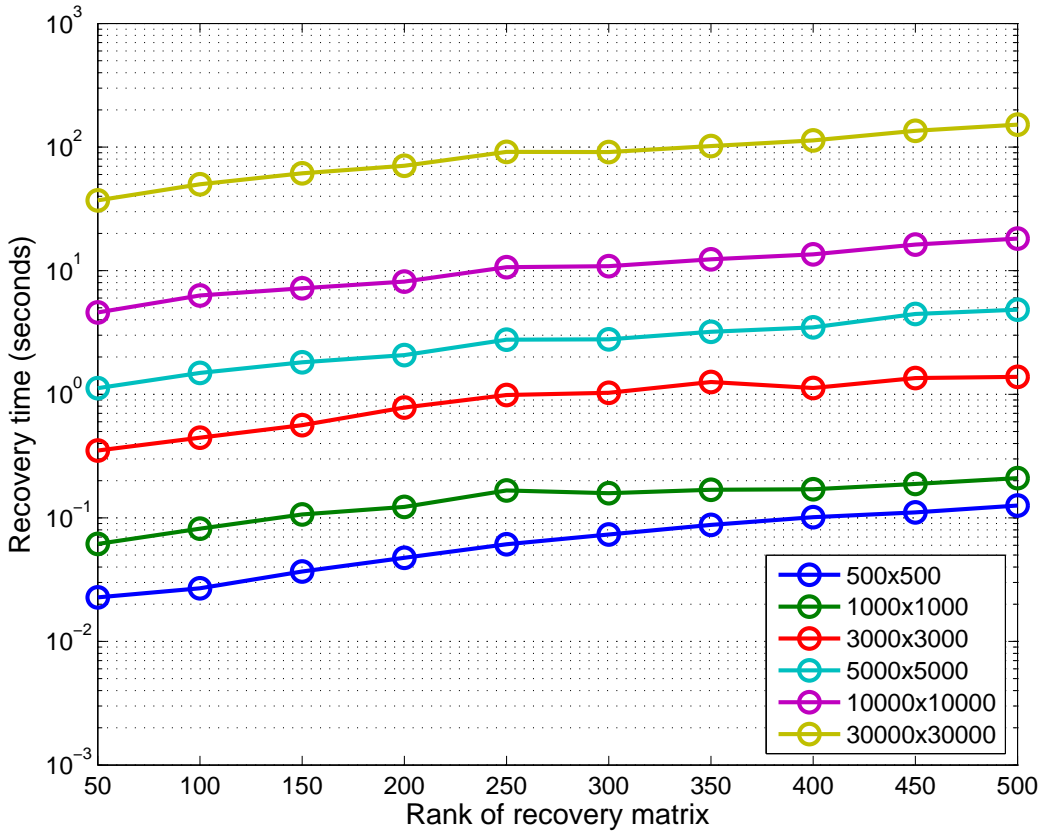}
\end{center}
   \caption{low-rank matrix recovery via BRP: the recovery time for matrices of different size and different rank.}
\label{fig:ranktime}
\end{figure}

We first evaluate the efficiency of the BRP based low-rank approximation (\ref{E:lr_app}) for exact recovery of low-rank matrices. We consider square matrices of dimension $n$ from $500$ to $30000$ with rank $r$ from $50$ to $500$. Each matrix is generated by $AB$, wherein $A$ and $B$ are both $n\times r$ standard Gaussian matrices. Figure \ref{fig:ranktime} shows that the recovery time is linearly increased w.r.t $n$. This is consistent with the $r^2(2n+r)+mnr$ flops required by (\ref{E:lr_app}). The relative error of each recovery is less than $10^{-14}$. It also shows that a $30000\times 30000$ matrix with rank $500$ can be exactly recovered within $200$ CPU seconds. This suggests the advantage of (\ref{E:lr_app}) for large-scale applications.

\begin{figure}[htb]
\begin{center}
 \includegraphics[width=1\linewidth]{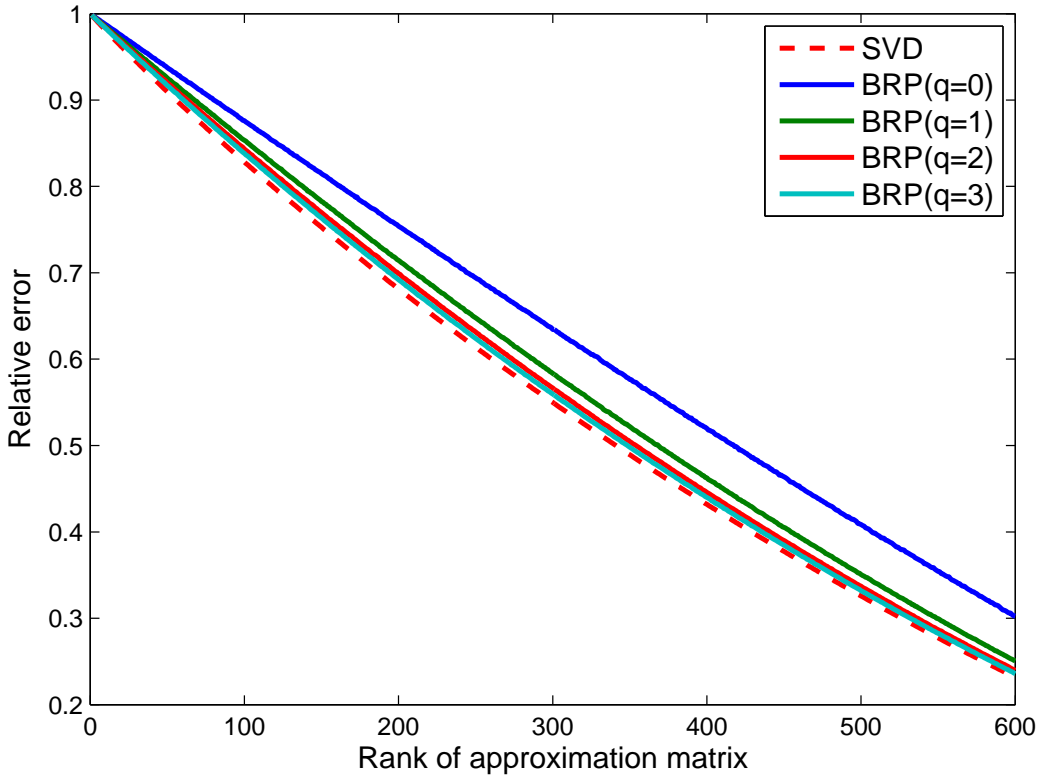}
\end{center}
   \caption{low-rank approximation via BRP: the relative approximation error for a $1000\times 1000$ matrix with standard normal distributed entries on different rank.}
\label{fig:rankerror}
\end{figure}

We then evaluate the effectiveness of (\ref{E:lr_app}) and its power scheme modification (\ref{E:mlr_app}) in low-rank approximation of full rank matrix with slowly decaying singular values. We generate a square matrix with size $1000$, whose entries are independently sampled from a standard normal distribution with mean $0$ and variance $1$, and then apply (\ref{E:lr_app}) ($q=0$) and (\ref{E:mlr_app}) with $q=1,2,3$ to obtain approximations with rank varying from $1$ to $600$. We show the relative errors in Figure \ref{fig:rankerror} and the relative error of the corresponding SVD approximation as a baseline. The results suggest that our method can obtain a nearly optimal approximation when $q$ is sufficiently large (e.g., 2).

\begin{figure}[htb]
\begin{center}
 \includegraphics[width=1\linewidth]{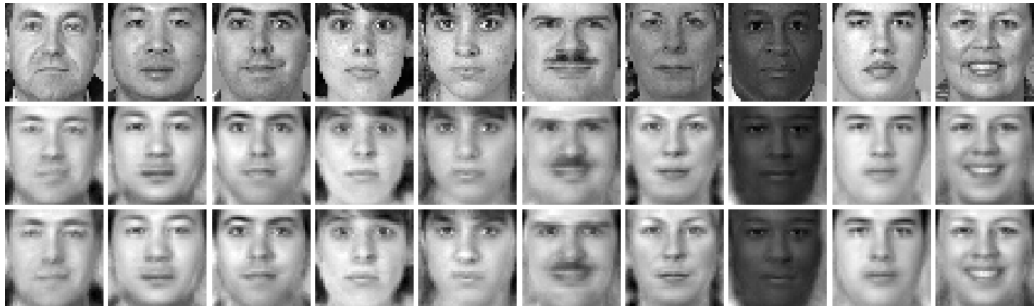}
\end{center}
   \caption{low-rank image compression via BRP on FERET: BRP compresses $700$ $40\times 40$ face images sampled from $100$ individuals to a $700\times 1600$ matrix with rank $60$. Upper row: Original images. Middle row: images compressed by SVD (6.59s). Bottom row: images compressed by BRP (0.36s).}
\label{fig:face}
\end{figure}

At last, we evaluate the efficiency and effectiveness of BRP on low-rank compression of human face images from dataset FERET \cite{FERET}. We randomly selected $700$ face images of $100$ individuals from FERET and built a $700\times 1600$ data matrix, wherein the $1600$ features are the $40\times 40$ pixels of each image. We then obtain two rank-$60$ compressions of the data matrix by using SVD and the power modification of BRP based low-rank approximation (\ref{E:mlr_app}) with $q=1$, respectively. The compressed images and the corresponding time costs are shown in Figure \ref{fig:face} and its caption. It indicates that our method is able to produce compression with competitive quality in considerably less time than SVD.

\section{Conclusion}

In this paper, we consider the problem of fast low-rank approximation. A closed form solution for low-rank matrix approximation from bilateral random projections (BRP) is introduced. Given an $m\times n$ dense matrix, the approximation can be calculated from $(m+n)r$ random measurements in $r^2(2n+r)+mnr$ flops. Power scheme is applied for improving the approximation precision of matrices with slowly decaying singular values. We prove the BRP based low-rank approximation is nearly optimal. The experiments on both artificial and real datasets verifies the effectiveness and efficiency of BRP in both low-rank matrix recovery and approximation tasks.

\newpage
\bibliography{BRPref}
\bibliographystyle{plain}

\end{document}